\documentclass[10pt,twocolumn,letterpaper]{article}

\usepackage{cvpr}
\usepackage{times}
\usepackage{epsfig}
\usepackage{graphicx}
\usepackage{amsmath}
\usepackage{amssymb}
\usepackage{bbm}
\usepackage{multirow}
\usepackage{float}
\usepackage[pagebackref=true,breaklinks=true,colorlinks,bookmarks=false]{hyperref}

\cvprfinalcopy % *** Uncomment this line for the final submission

\begin{document}

%%%%%%%%% TITLE
\title{Bilateral Network with Channel Splitting Network and Transformer for Thermal Image Super-Resolution}

% \author{Bo Yan, Leilei Cao, Fengliang Qi, Hongbin Wang\\
% Ant Group \\
% }

\author{Bo Yan, Leilei Cao, Fengliang Qi, Hongbin Wang \\
Ant Group, China \\
{\{lengyu.yb, caoleilei.cll, qifengliang.qfl, hongbin.whb\}@antgroup.com} \\
}

\maketitle
%\maketitle
%\thispagestyle{empty}

%%%%%%%%% ABSTRACT
\begin{abstract}
In recent years, the Thermal Image Super-Resolution (TISR) problem has become an attractive research topic. TISR would been used in a wide range of fields, including military, medical, agricultural and animal ecology. Due to the success of PBVS-2020 and PBVS-2021 workshop challenge, the result of TISR keeps improving and attracts more researchers to sign up for PBVS-2022 challenge. In this paper, we will introduce the technical details of our submission to PBVS-2022 challenge designing a Bilateral Network with Channel Splitting Network and Transformer(BN-CSNT) to tackle the TISR problem. Firstly, we designed a context branch based on channel splitting network with transformer to obtain sufficient context information. Secondly, we designed a spatial branch with shallow transformer to extract low level features which can preserve the spatial information. Finally, for the context branch in order to fuse the features from channel splitting network and transformer, we proposed an attention refinement module, and then features from context branch and spatial branch are fused by proposed feature fusion module. The proposed method can achieve PSNR=33.64, SSIM=0.9263 for x4 and PSNR=21.08, SSIM=0.7803 for x2 in the PBVS-2022 challenge test dataset.

\end{abstract}

%%%%%%%%% BODY TEXT
\section{Introduction}

In recent years, the Thermal Image Super-Resolution (TISR) problem has become a challenging and great significance task would been widely used in military, medical, agricultural and animal ecology fields. PBVS-2022 challenge stated consists in obtaining super-resolution images at x2 and x4 scales from the given images.

Due to the success of PBVS-2020\cite{Rivadeneira_2020_CVPR_Workshops} and PBVS-2021\cite{Rivadeneira_2021_CVPR} challenge, a lot of effective models have been proposed for TISR. Channel Splitting Network(CSN)\cite{zhao2019channel, prajapati2021channel} is widely used for this task and gets an impressive result. Transformers have made enormous strides in NLP\cite{devlin2018bert, radford2019language}. There are quite a bit of works applying transformers to computer vision\cite{ramachandran2019stand, Zhao_2020_CVPR, carion2020end} because transformers can capture the non-local and relational nature of images. Especially, Swin Transformer\cite{liu2021swin} has been widely used for many computer vision tasks and  achieves successful results.

Inspired by Channel Splitting Network and Swin Transformer, we design a Bilateral Network with Channel Splitting Network and Transformer. The context branch based on channel splitting network with transformer can obtain sufficient context information. And the spatial branch with shallow transformer can preserve the spatial information. Finally, the attention refinement module and feature fusion module are designed to fuse features. The experiments show that the proposed method can achieve a competitive performance on  PBVS-2022 challenge test dataset.
%-------------------------------------------------------------------------
\section{Proposed Methodology}

\begin{figure*}[ht]
	\centering
	\includegraphics[scale=0.58]{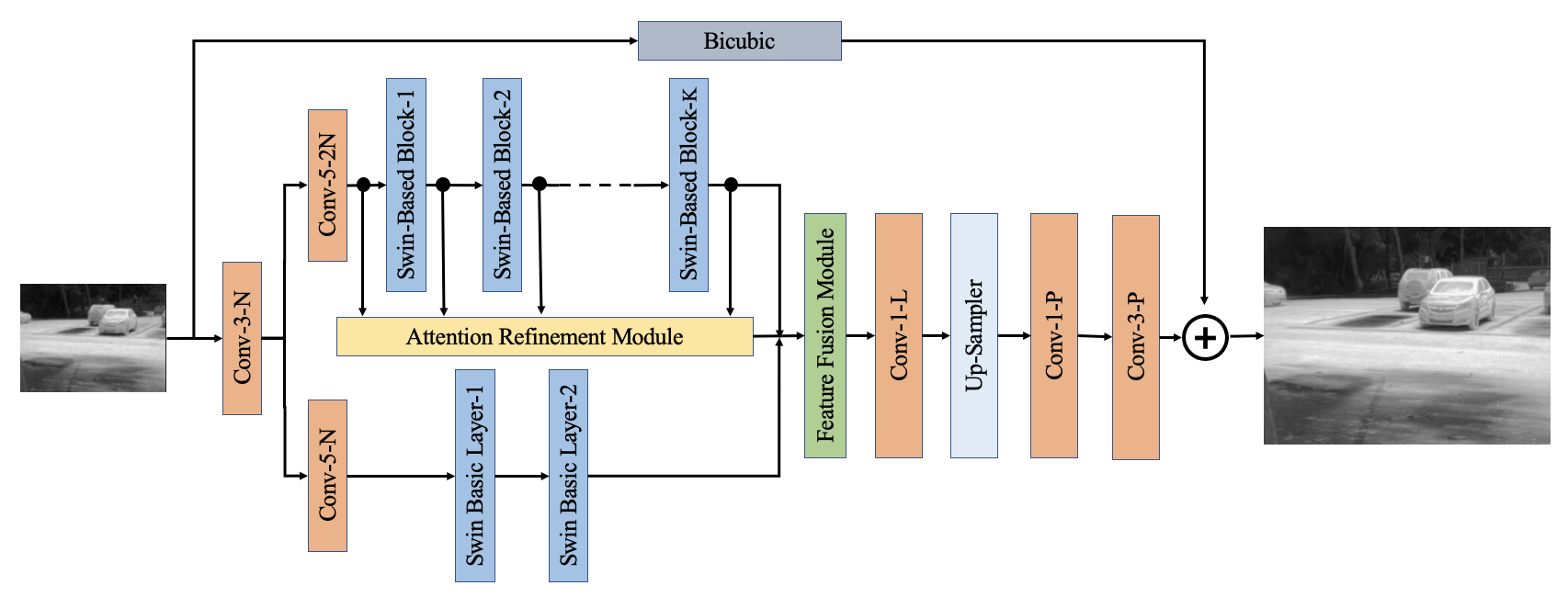}
	\caption{The proposed BN-CSNT framework for up-scaling ×4 and up-scaling ×2 of PBVS-2022 Thermal SR Challenge.}
	\label{pipeline}
\end{figure*}

\begin{figure*}[ht]
	\centering
	\includegraphics[scale=0.58]{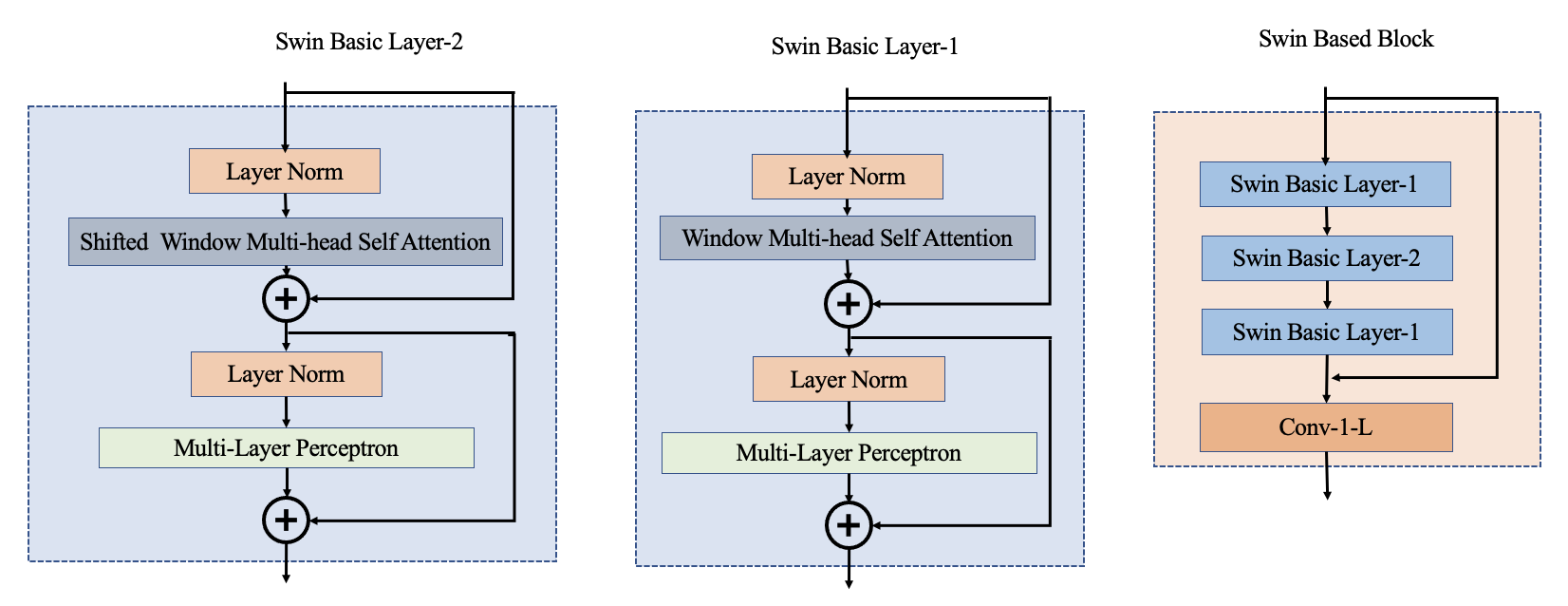}
	\caption{The architecture design of Swin-Based Block of the proposed network.}
	\label{pipeline}
\end{figure*}

\begin{figure*}[ht]
	\centering
	\includegraphics[scale=0.58]{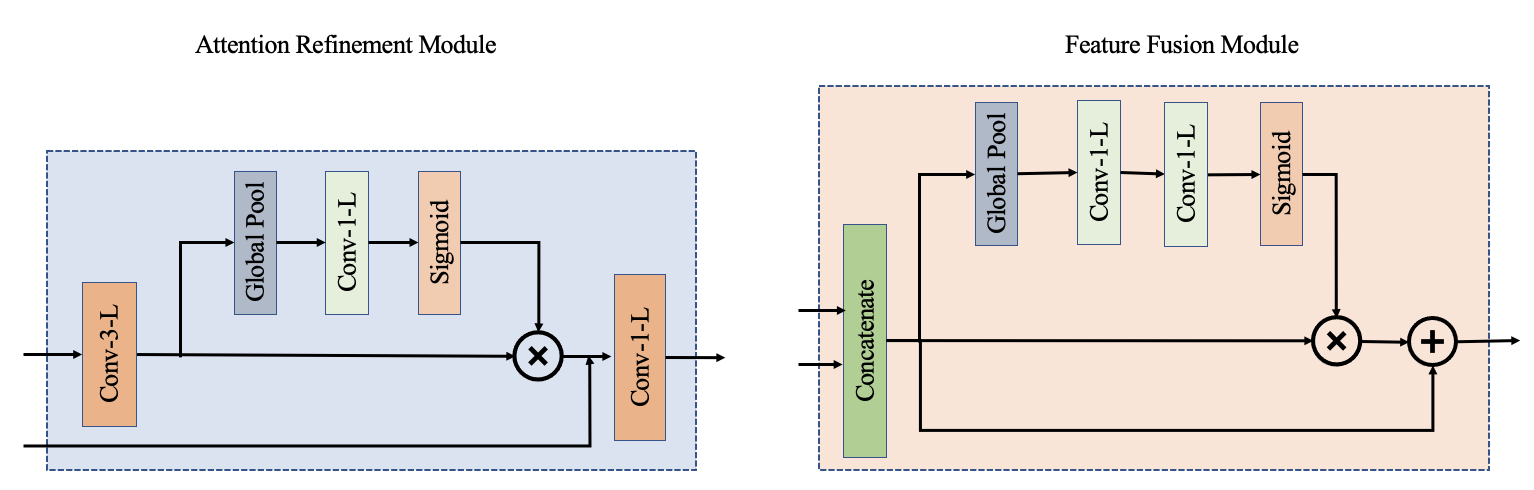}
	\caption{The architecture design of Attention Refinement Module and Feature Fusion Module.}
	\label{pipeline}
\end{figure*}

We design a Bilateral Network with Channel Splitting Network and Transformer, the structure is illustrated in Figure 1. 

Firstly, the input low resolution image is passed through the convolution layer with a kernel size of 3 × 3 and N channels and then input to context branch and spatial branch.

The context branch based on channel splitting network with transformer can obtain sufficient context information, and it is composed of K(K=8) Swin-Based Blocks. The details of Swin-Based Block are illustrated in Figure 2. For context branch, the input feature maps are passed through the convolution layer with a kernel size of 5 × 5 and 2N channels, then split N feature-maps input to Swin-Based Block and another N feature-maps input to Attention Refinement Module(ARM). Each Swin-Based Block unit takes N channel features and outputs 2N number of channels. The input N feature-maps to Swin-Based Block are passed through the Swin Basic Layers and output N feature-maps, then N input feature-maps and N output  feature-maps are concatenated and passed through the convolution layer with a kernel size of 1 × 1 and 2N channels. For the output 2N feature-maps of Swin-Based Block, split N feature-maps to next Swin-Based Block and another N feature-maps to ARM, the design of ARM is depicted in Figure 3.

For spatial branch, the input feature maps are passed through the convolution layer with a kernel size of 5 × 5 and N channels, and then passed through shallow(2) Swin Basic Layers to preserve the spatial information.

The features obtained from context branch, ARM and spatial branch are concatenated and fused by Feature Fusion Module(FFM), the design of FFM is depicted in Figure 3. Finally, we employ the pixel-shuffle operator to increase the spatial resolution of the feature\cite{ shi2016real}.

\section{Experimental Analysis}
The ground truth images for the x4 scale correspond to the provided high-resolution images, so we down-sample the given images by x4 adding noise and use these down-sampled images as inputs to develop x4 super-resolution solutions. 

Regarding the x2 super-resolution solution, it should be developed using as an input the given mid-resolution images acquired with a camera and as an output, the corresponding high-resolution images, of the same scene, but acquired with another camera. So the x2 scale proposed solution should be able to tackle both problems generating the super-resolution of the images acquired with the camera to another camera.

\subsection{Training and Testing datasets}
In Track-1, the ground truth images are Flir dataset with high resolution, the low resolution images are generated by bicubic down-sampling with a factor ×4 and degrading the same by Additive White Gaussian Noise (AWGN) with mean and standard deviation values of 0 and 10.

For Track-2, the target is required with an upscaling factor of ×2 for Axis dataset with medium resolution, and match the high resolution of Flir dataset with same scene. The available images from Axis and Flir dataset are not exactly registered with pixel-wise accuracy. Hence, we perform image registration using SIFT features\cite{lindeberg2012scale} which results in semi-matched image pairs. And also, we  down-sample the medium resolution Axis dataset with x2 by bicubic downsampling to  get the low resolution Axis dataset.

\subsection{Training Strategy}
For Track-1, we train the BN-CSNT network with an upscaling factor of ×4 with L1 loss, and the inputs are down-sampling with a factor ×4 and degrading Flir images and outputs are high resolution Flir images.

For Track-2, we firstly train the BN-CSNT network with an upscaling factor of ×2 with L1 loss, and the inputs are down-sampling with a factor ×2 Axis images and outputs are medium resolution Axis images. Secondly, we train the BN-CSNT network with an upscaling factor of ×2 with L1 loss, Least Squared GAN(LSGAN) loss\cite{mao2017least} and SSIM loss, and the inputs are semi-matched medium resolution Axis images and outputs are high resolution Flir images. The average of outputs of these two models are the final result for track-2.

\subsection{Experimental Results}
The Peak Signal-to-Noise Ratio (PSNR) and Structural Similarity Index Measure (SSIM) metrics are used to validate the SR performance. As shown in Table 1, Our proposed method finally achieves PSNR=33.64, SSIM=0.9263 for x4 and PSNR=21.08, SSIM=0.7803 for x2 in the PBVS-2022 challenge test dataset.

\begin{table}[ht]
\centering
\caption{Experimental results on PBVS-2022 test set}
\label{Quantitative3}
\vspace{0.1in}
\begin{tabular}{lcc}
\hline
Track & PSNR & SSIM \\
\hline
PBVS-2022 Track-1 x4 & 33.64 & 0.9263 \\
PBVS-2022 Track-2 x2 & 21.08 & 0.7803 \\
\hline
\end{tabular}
\end{table}

%------------------------------------------------------------------------
\section{Conclusions}
In this paper, we introduce the technical details of our submission to the PBVS-2022 challenge. We propose a Bilateral Network with Channel Splitting Network and Transformer for thermal image super-resolution, including a context branch with channel splitting network and transformer, a spatial branch with shallow transformer, and an attention refinement module and feature fusion module, and also a training strategy. Finally, Our approach achieves PSNR=33.64, SSIM=0.9263 for x4 and PSNR=21.08, SSIM=0.7803 for x2 in the PBVS-2022 challenge test.

%{\small
%\bibliographystyle{ieee_fullname}
%\bibliography{egpaper_final}
%}
%\input{egpaper_final.bbl}

%%%%%%%%% REFERENCES
{\small
\bibliographystyle{ieee_fullname}
\bibliography{egbib}
}

\end{document}